# DRO: Deep Recurrent Optimizer for Structure-from-Motion

Xiaodong Gu[1*]   Weihao Yuan[1*]   Zuozhuo Dai[1]   Chengzhou Tang[2]   Siyu Zhu[1]   Ping Tan[12]
[1]Alibaba A.I. Labs    [2]Simon Fraser University

## Abstract

*There are increasing interests of studying the structure-from-motion (SfM) problem with machine learning techniques. While earlier methods directly learn a mapping from images to depth maps and camera poses, more recent works enforce multi-view geometry constraints through optimization embedded in the learning framework. This paper presents a novel optimization method based on recurrent neural networks to further exploit the potential of neural networks in SfM. Specifically, our neural optimizer alternately updates the depth and camera poses through iterations to minimize a feature-metric cost, and two gated recurrent units iteratively improve the results by tracing historical information. In this way, our network is a gradient-free zeroth-order optimizer designed for SfM and can be applied to both supervised and self-supervised SfM. Extensive experimental results demonstrate that our method outperforms previous methods and is more efficient in computation and memory consumption than cost-volume-based methods. In particular, our self-supervised method outperforms previous supervised methods on the KITTI and ScanNet datasets. Our source code is available at* https://github.com/aliyun/dro-sfm.

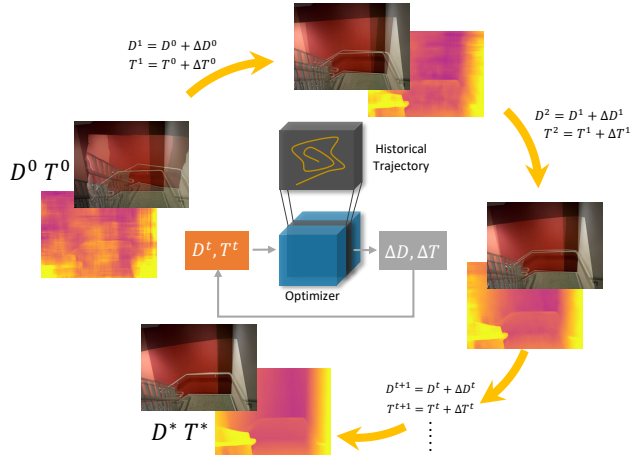

Figure 1: A gated recurrent network (indicated as the optimizer in the middle) iteratively updates a depth map and the relative motion between two images by minimizing a feature-metric cost. At each iteration, we show a color-coded depth map and a superimposed image generated from the two input images according to the depth map and camera motion. Over the iterations, the superimposed image becomes gradually sharper while the depth map improves.

## 1. Introduction

Structure-from-motion (SfM) [31] is a fundamental task in computer vision and essential for numerous applications such as robotics, autonomous driving, augmented reality, and 3D reconstruction. Given a sequence of images, SfM methods optimize depth maps and camera poses to recover the 3D structure of a scene. Traditional methods solve the Bundle-Adjustment (BA) problem [36], where the reprojection error between reprojected 3D scene points and 2D image feature points are minimized iteratively.

Recently, deep-learning-based methods have dominated most benchmarks and demonstrated advantages over traditional methods [18, 22, 33, 34, 37, 41]. Earlier learning-based methods [14, 24, 27, 37] directly regress the depth maps and camera poses from the input images, but the domain knowledge such as multi-view geometry is ignored.

To combine the strength of neural networks and traditional geometric methods, more recent works formulate the geometric-based optimization as differentiable layers and embed them in a learning framework [33, 34, 46].

We follow the approach of combining neural networks and optimization methods with some novel insights. Firstly, previous methods [12, 33, 34] adopt gradient-based optimization such as Levenberg-Marquardt or Gauss-Newton methods. However, the gradients could be noisy and misleading especially for the high-dimensional optimization problem in dense depth map computation. Traditional methods like LSD-SLAM [16] and DSO [15] focus on edge pixels to make the optimization problem feasible. More recent learning-based methods carefully design learned regularization such as depth bases [33] or manifold embeddings [6, 7] to address this problem. However, this learned regularization might have difficulty to be generalized to unseen scenes. Furthermore, a multi-resolution strategy is needed to gradually compute the solution from coarse to fine. In comparison, we employ a gated recurrent neural network for optimization as inspired by [35]. An illustration is shown in Figure 1. Remarkably, our method is gradient-free and works on the high resolution image directly without any regularization which might limit the algorithm generalization.

---

[*]Equal contribution.

Secondly, some methods [34, 41, 46, 49] build cost volumes to solve dense depth maps. Similar cost volumes are also employed in [35] to compute optical flow. A cost volume encodes the errors of multiple different depth values at each pixel. It evaluates the result quality within a large spatial neighborhood in the solution space in a discrete fashion. While cost volumes have been demonstrated effective in computing depth maps [21, 41, 44], they are inefficient in time and space because they exhaustively evaluate results in a large spatial neighborhood. We argue that a gated recurrent network [11] can minimize the feature-metric error to compute dense depth without resorting to compute such a cost volume. A gated recurrent network updates depth maps only by evaluating the current solution (i.e. a single point in the solution space) and some previous solutions over time. In spirit, our learned recurrent network is a zeroth-order optimization method and exploits temporal information during iterations, while gradient based methods or cost volume based methods rely only on spatial information. In this way, our method has the potential of better running time and memory efficiency.

In experiments, we show that our recurrent optimizer reduces the feature-metric cost over iterations and produces gradually improved depth maps and camera poses. Our method demonstrates better accuracy than previous methods in both indoor and outdoor data, under both supervised and self-supervised settings. In particular, our self-supervised method outperforms most previous supervised methods on both KITTI and ScanNet datasets.

## 2. Related work

**Supervised Deep SfM.** Deep neural networks can learn to solve the SfM problem directly from data [37, 49, 41]. With the ground-truth information, DeMoN [37] trains two network branches to regress structures and motions separately with an auxiliary flow prediction task to exploit feature correspondences. Some methods adopt a discrete sampling strategy to achieve high-quality depth maps [49, 34]. They generate depth hypotheses and utilize multiple images to construct a cost volume. Furthermore, the pose volume is also introduced in [41]. They take the feature maps to build two cost volumes and employ 3D convolutions to regularize. There are also methods to directly regress scene depth from a single input image [14, 18, 27], which is an ill-posed problem. These methods rely heavily on the data fitting of the neural networks. Therefore, their network structure and feature volumes are usually bulky, and their performance are limited in unseen scenes.

**Self-supervised Deep SfM.** Supervised methods, nevertheless, require collecting a large number of training data with ground-truth depth and camera poses. Recently, many self-supervised works [4, 19, 22, 28, 29, 30, 38, 39, 43, 45, 47, 50] have been proposed to train a depth and pose estimation model from only monocular RGB images. They employ the predicted depths and poses to warp neighbor frames to the reference frame, such that a photometric constraint is created to serve as a self-supervision signal. In this case, the dynamic objects is a problem and would generate errors in the photometric loss. To address this, semantic mask [25] and optical flow [51, 48, 5] are proposed to exclude the influence of moving objects. Another challenge is the visibility between different frames. To deal with this, a minimum re-projection loss is designed in [19, 22] to handle the occluded regions. Despite these efforts, there is still a gap between self-supervised and supervised methods.

**Learning to Optimize.** Traditional computer vision methods usually formulate the tasks as optimization problems according to the first principles such as photo-consistency and multi-view geometry. Many of recent works try to combine the strength of neural networks and traditional optimization-based methods. There are mainly two approaches in learning to optimize. One approach [2, 3, 33, 34] employs a network to predict the inputs or parameters of an optimizer, which is implemented as some layers in a large neural network for end-to-end training. On the contrary, the other approach directly learn to update optimization variables from the data [1, 9, 12, 17, 35]. There, however, are some problems in previous methods. Methods of the first approach need to explicitly formulate the solver and are limited to problems where the objective functions are easily defined [3, 2, 33, 34]. Furthermore, the methods in [12, 33] need to explicitly evaluate gradients of the objective function, which is hard in many problems. Besides, the methods in [34, 35] adopt cost volumes, which make the model heavy to apply.

In comparison, our method does not require gradients computation or cost volume aggregation. It only evaluates the result quality at a single point in the solution space at each step. In this sense, it is a zeroth-order optimizer embedded in a network. By accumulating temporal evidence from previous iterations, our GRU module learns to minimize the objective function. Unlike the method in [35] which still relies on a cost volume, our method is more computation and memory efficient. Besides, two updaters in our framework, one for depth and the other one for pose, are alternately updated, which is inspired by the traditional bundle adjustment algorithm.

## 3. Deep Recurrent Optimizer

### 3.1. Overview

Given a reference image $\mathbf{I}_0$ and $N$ neighboring images $\{\mathbf{I}_i\}_{i=1}^{N}$, our method outputs the depth $\mathbf{D}$ of the reference image and the relative camera poses $\{\mathbf{T}_i\}_{i=1}^N$ for images $\{\mathbf{I}_i\}_{i=1}^N$ as shown in Figure 2. Images first go through a shared feature extraction module to produce feature maps $\mathcal{F}_i$. A depth head and a pose head then take in these features and output an initial depth map $\mathbf{D}^0$ and initial relative poses $\mathbf{T}_i^0$. Finally, these initial results are iteratively refined by the depth and the pose GRU-optimizers alternately, and converge to the final depth $\mathbf{D}^*$ and poses $\mathbf{T}_i^*$.

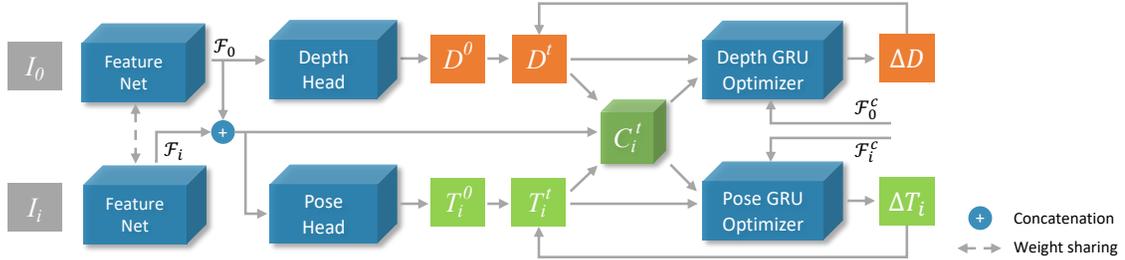

Figure 2: The overview of our framework. The reference image $\mathbf{I}_0$ and context images $\mathbf{I}_i$ are first fed into a feature network with shared parameters to extract feature maps. The extracted features are then mapped to an initial depth map $\mathbf{D}^0$ and an initial relative camera pose $\mathbf{T}_i^0$ by the depth head and the pose head, respectively. Afterwards, the recurrent optimizer update the depth and pose iteratively. During this optimization, we first compute a cost map $\mathbf{C}_i^t$ from the current solution $\mathbf{D}^t$ and $\mathbf{T}_i^t$. The GRU optimizer then computes the solution updates $\Delta \mathbf{D}$ and $\Delta \mathbf{T}_i$ accordingly. Eventually, the depth map and pose gradually converge to the optimum, $\mathbf{D}^*$ and $\mathbf{T}_i^*$.

### 3.2. Feature Extraction and Cost Construction

**Feature extraction.** We use ResNet18 [23] as the backbone to extract the features $\{\mathcal{F}_i\}_{i=0}^N$. The resolution of the feature maps is $1/8$ of the original input images. Likewise, the initial hidden state defined by $h^0$ and the contextual features $\{\mathcal{F}_i^c\}_{i=0}^N$ for the GRU optimizer come from the same-structure feature network.

**Cost Construction** Similar to BA-Net [33], we construct a photometric cost in feature space as the energy function to minimize. The cost measures the distance between aligned feature maps. Given the depth map $\mathbf{D}$ of the reference image $\mathbf{I}_0$ and the relative camera pose $\mathbf{T}_i$ of another image $\mathbf{I}_i$ with respect to $\mathbf{I}_0$, the cost is defined at each pixel $x$ in the reference image $\mathbf{I}_0$:

$$\mathbf{C}_i(x) = ||\mathcal{F}_i(\pi(\mathbf{T}_i \circ \pi^{-1}(x, \mathbf{D}(x)))) - \mathcal{F}_0(x)||_2, \quad (1)$$

where $||\cdot||_2$ is the L2 norm, and $\pi(\cdot)$ is the projection of a 3D point to the image plane. Thus, its inverse function $\pi^{-1}(x, \mathbf{D}(x))$ maps a pixel $x$ and its depth $\mathbf{D}(x)$ back to a 3D point. The transformation $\mathbf{T}_i$ convert 3D points from the camera space of $\mathbf{I}_0$ to that of $\mathbf{I}_i$. When there are multiple neighboring images, we average multiple cost values $\{\mathbf{C}_i(x)\}_{i=1}^N$ as $\mathbf{C}(x)$ at each pixel:

$$\mathbf{C}(x) = \frac{1}{N}\sum_{i=1}^N \mathbf{C}_i(x). \quad (2)$$

Note that the feature-metric error in BA-Net [33] will further sum the cost over all pixels as $\sum_x \mathbf{C}(x)$. However,

in this work, we maintain a cost map $\mathbf{C}(x)$ with the same resolution as the feature map $\mathcal{F}_i$ to facilitate minimization of the feature-metric error. In the following of this paper, we refer $\mathbf{C}(x)$ as cost map instead of feature-metric error. In alternate optimization, the camera pose $\mathbf{T}_i$ is optimized by minimizing $\mathbf{C}_i(x)$ and the depth map $\mathbf{D}$ is optimized by minimizing the averaged cost $\mathbf{C}(x)$.

### 3.3. Iterative Optimization

We minimize the cost map $\mathbf{C}$ in an iterative manner. At each iteration, the optimizer updates the results as

$$\mathbf{D}^{t+1} \leftarrow \mathbf{D}^t + \Delta \mathbf{D}^t, \qquad \mathbf{T}_i^{t+1} \leftarrow \mathbf{T}_i^t + \Delta \mathbf{T}_i^t, \quad (3)$$

where $\Delta \mathbf{D}$ and $\Delta \mathbf{T}_i$ are the updates. Inspired by [35], we utilize a gated recurrent unit to compute these updates, since a GRU can memorize the status of previous steps and can effectively exploit temporal information during the optimization process.

#### 3.3.1 Initialization

The depth map and relative poses are initialized by two different heads on top of the feature extraction network. The depth head consists of two convolution layers, while the pose head is further equipped with an additional average pooling layer. The hidden state of the GRU is initialized by the contextual features with the tanh function as activation.

#### 3.3.2 Recurrent Update

We design two GRU modules, one for updating the depth and the other one for updating the camera pose. Each GRU module receives the current cost map $\mathbf{C}^t$ and the current estimated variables $\mathbf{V}^t$ (depth map $\mathbf{D}^t$ or camera pose $\mathbf{T}^t$) and outputs an incremental update $\Delta \mathbf{V}^t$ to update the results as $\mathbf{V}^{t+1} = \mathbf{V}^t + \Delta \mathbf{V}^t$.

Specifically, we first project the variable $\mathbf{V}^t$ and the cost map $\mathbf{C}^t$ into the feature space with $\mathcal{P}_v$ and $\mathcal{P}_c$, both of which are composed of two convolutional layers. We then concatenate $\mathcal{P}_v(\mathbf{V}^t)$, $\mathcal{P}_c(\mathbf{C}^t)$, and the image contextual feature $\mathcal{F}^c$ to form a tensor $\mathbf{M}^t$ as the input. The structure inside each GRU unit is as follows:

$$\begin{aligned}
z^{t+1} &= \sigma(\text{Conv}_{5\times 5}([h^t, \mathbf{M}^t], W_z)) \\
r^{t+1} &= \sigma(\text{Conv}_{5\times 5}([h^t, \mathbf{M}^t], W_r)) \\
\tilde{h}^{t+1} &= \tanh(\text{Conv}_{5\times 5}([r^{t+1} \odot h^t, \mathbf{M}^t], W_h)) \\
h^{t+1} &= (1 - z^{t+1}) \odot h^t + z^{t+1} \odot \tilde{h}^{t+1},
\end{aligned} \quad (4)$$

where $\text{Conv}_{5\times 5}$ represents a separable $5 \times 5$ convolution, $\odot$ is the element-wise multiplication, $\sigma$ and $\tanh$ are the sigmoid and the tanh activation functions. Finally, the depth maps or the camera poses are predicted from the hidden state $h^t$ by the same structures as the initial depth or pose head in Sec. 3.3.1.

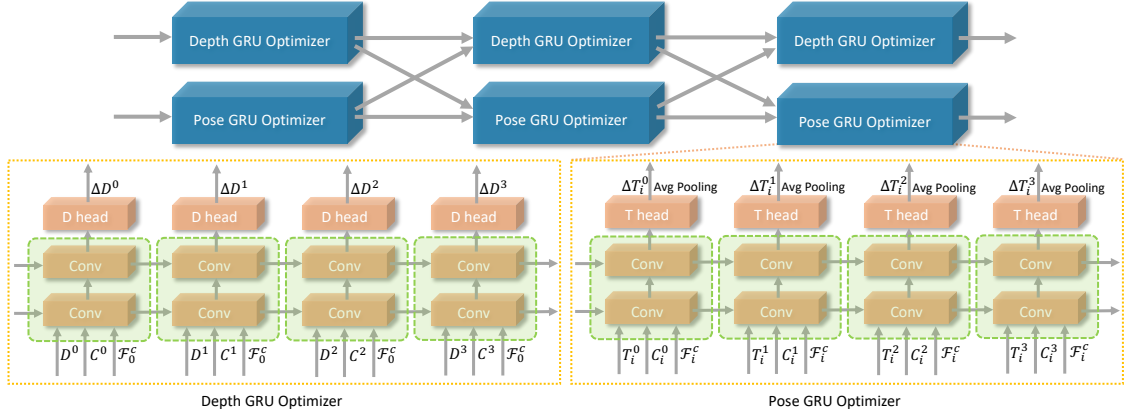

Figure 3: Working flow of the optimizer. Updating of the depth and the pose are separated in each stage, where 4 updates of the depth are followed by 4 updates of the pose. We adopt 3 stages in our framework. For each update for the depth, the predicted depth $\mathbf{D}^t$, cost $\mathbf{C}^t$, and contextual feature map $\mathcal{F}_0^c$ are fed in, then the update $\Delta \mathbf{D}^t$ is predicted based on the inputs and historical information. Afterwards, the depth is updated by $\mathbf{D}^{t+1} = \mathbf{D}^t + \Delta \mathbf{D}^t$.

With this optimizer, starting from an initial solution, the estimated depths and poses are iteratively refined by the optimization iterations and eventually converge to the final results as, $\mathbf{D}^t \to \mathbf{D}^*$ and $\mathbf{T}_i^t \to \mathbf{T}_i^*$.

#### 3.3.3 Alternate Optimization

After defining the structure of the GRU module, we update the depth map $\mathbf{D}^t$ and camera poses $\mathbf{T}_i^t$ alternately with $m$ stages. As shown in Figure 3, at each stage, we first freeze the camera pose and update the depth map as $\mathbf{D}^{t+1} = \mathbf{D}^t + \Delta \mathbf{D}^t$, which is repeated by $n$ times. We then freeze the depth map $\mathbf{D}$ and update the camera poses with $\mathbf{T}_i^{t+1} = \mathbf{T}_i^t + \Delta \mathbf{T}^t$, which is also repeated by $n$ times. This alternative optimization leads to more stable optimization and easier training empirically. In experiments, we fix $m$ at 3 and $n$ at 4, unless specified otherwise.

To gain more insights into the recurrent process and demonstrate the GRU module behaves as a recurrent optimizer, we visualize how the feature-metric error decreases over the GRU iterations on the first row of Figure 4. The blue curves are the plots of the feature-metric cost over iterations, while the yellow curves are the errors in depth map and relative camera poses compared with a known ground truth. In general, the feature-metric error decreases, but not strictly monotonically decreases. The error might increase and then decrease again, indicating our optimizer has the capability of jumping out of local minimums.

The last two rows of Figure 4 further visualize the intermediate results over the iterations. At the bottom are the two input images and the ground truth depth map. In the middle, shown are the cost map $\mathbf{C}$, estimated depth map, and a warp-aligned image that is composed by warping the neighbor image according to the estimated depth and camera pose and superimpose with the reference image. From top to bottom we can see the warp-aligned image becomes sharper, indicating higher quality of estimated depth and camera pose.

### 3.4. Training Loss

Our method can be applied to both supervised and self-supervised SfM. In the following, we introduce the training loss for both schemes.

#### 3.4.1 Supervised Case

When the ground truth is available, we supervise the training by evaluating the depth and pose errors.

**Depth supervision** $\mathcal{L}_{\text{depth}}$ computes the L1 distance between the predicted depth map $\mathbf{D}$ and the ground-truth depth map $\hat{\mathbf{D}}$ in each stage:

$$\mathcal{L}_{\text{depth}} = \sum_{s=1}^{m} \gamma^{m-s} ||\mathbf{D}^s - \hat{\mathbf{D}}||_1, \quad (5)$$

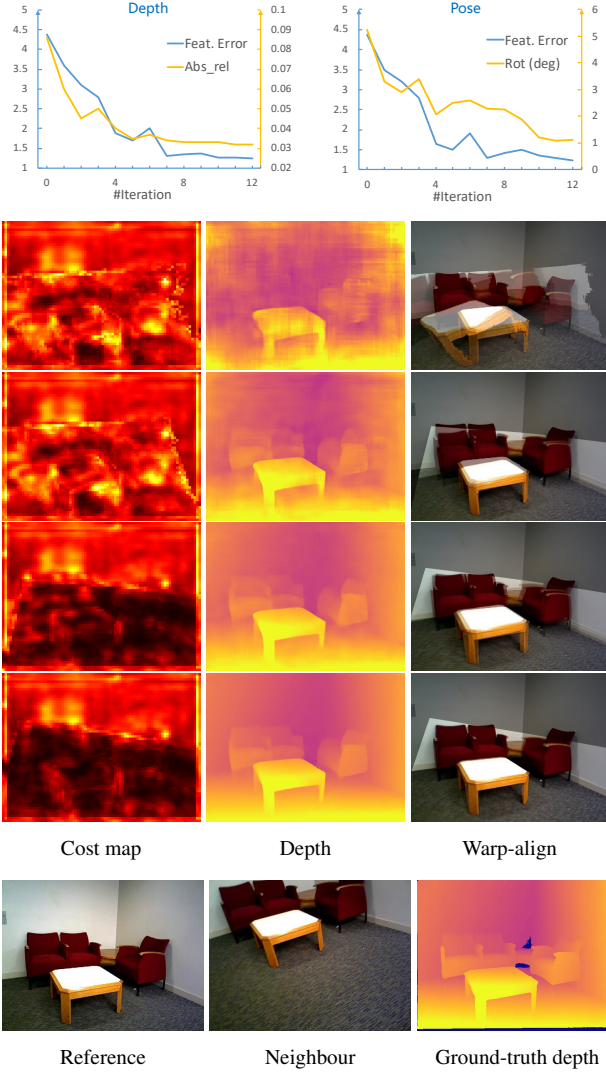

Figure 4: The first row shows the feature-metric error (the cost), depth error, and camera pose error decrease with regards to the iterations of the GRU optimizer. In the middle we show the estimated cost map, depth map, and a warp-aligned image at each iteration of the GRU optimizer. The warp-aligned image warps a neighbor image onto the reference image using the estimated depth and camera pose. It becomes clearer at later iterations, indicating a better result.

where $\gamma$ is a discounting factor and s is the stage number.

**Pose supervision** $\mathcal{L}_{\text{pose}}$ is defined as the following according to the ground truth depth $\hat{D}$ and pose $\hat{T}_i$:

$$\mathcal{L}_{\text{pose}} = \sum_{s=1}^{m} \sum_x \gamma^{m-s} ||\pi(T_i^s \circ \pi^{-1}(x, \hat{D}(x))) - \pi(\hat{T}_i \circ \pi^{-1}(x, \hat{D}(x)))||_1. \quad (6)$$

This loss computes the image re-projection of a pixel $x$ according to the estimated camera pose $T_i^s$ and the true pose $\hat{T}_i$ in each stage. The pose loss is the distance between these two projections, which is in the image space and insensitive to different scene scales. In the experiments we find it is more effective than directly comparing the difference between $T^s$ and $\hat{T}$.

The final supervised loss is the sum of these two terms:

$$\mathcal{L}_{\text{supervised}} = \mathcal{L}_{\text{depth}} + \mathcal{L}_{\text{pose}}. \quad (7)$$

### 3.4.2 Self-supervised Case

When the ground truth is not available, we borrow the loss defined in [20] for self-supervised training. Specifically, the supervision signal comes from geometric constraints and consists of two terms, a photometric loss and a smoothness loss.

**Photometric loss** $\mathcal{L}_{\text{photo}}$ measures the similarity of the reconstructed images $\{I'_i\}_{i=1}^N$ to the reference image $I_0$. Here, the reconstructed images $I'_i$ are generated by warping the input image $I_i$ according to the depth $D$ and pose $T_i$ to reconstruct $I_0$. This similarity is measured by the structural similarity (SSIM) [40] with L1 loss as

$$\mathcal{L}_{\text{photo}} = \alpha \frac{1 - SSIM(I'_i, I_0)}{2} + (1-\alpha)||I'_i - I_0||_1, \quad (8)$$

where $\alpha$ is a weighting factor. For the fusion of multiple photometric losses, we also take the strategies defined in [20], which adopts a minimum fusion and masks stationary pixels.

**Smoothness loss** $\mathcal{L}_{\text{smooth}}$ encourages adjacent pixels to have similar depths and is defined as:

$$\mathcal{L}_{\text{smooth}} = |\partial_x D|e^{-|\partial_x I_0|} + |\partial_y D|e^{-|\partial_y I_0|}. \quad (9)$$

Then the self-supervision loss is defined by a weighted sum of these two terms:

$$\mathcal{L}_{\text{self}} = \mathcal{L}_{\text{photo}} + \lambda \mathcal{L}_{\text{smooth}}, \quad (10)$$

where $\lambda$ enforces a weighted depth regularization.

## 4. Experiments

This section first presents the implementation details and then evaluates our supervised and self-supervised models on outdoor and indoor datasets, which are followed by the ablation studies.

### 4.1. Implementation Details

Our work is implemented in Pytorch experimented with on Nvidia GTX 2080 Ti GPUs. The network is optimized end-to-end with the Adam optimizer ($\beta_1 = 0.9$, $\beta_1 = 0.999$). The training runs for 50 epochs with the learning rate reduced from $2 \times 10^{-4}$ to $5 \times 10^{-5}$. For the supervised training, we use the losses described in Sec. 3.4.1 with $\gamma$ set as 0.85. For the self-supervised training, the losses are depicted in Sec. 3.4.2 with $\alpha$ set as 0.85 and $\lambda$ set as 0.01.

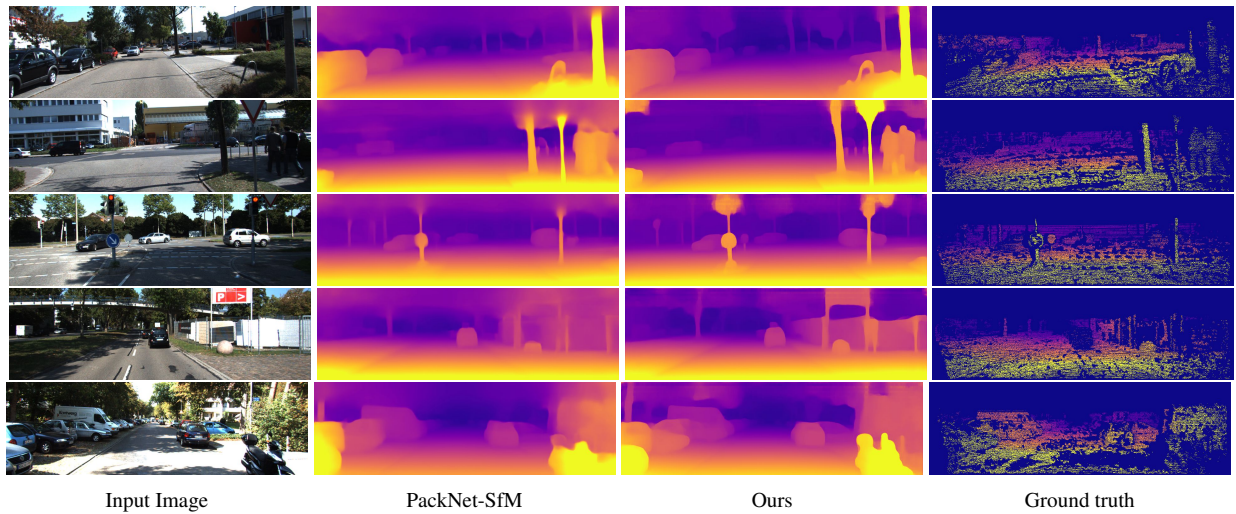

| | Input Image | PackNet-SfM | Ours | Ground truth |

Figure 5: Qualitative results on the KITTI dataset.

| Method | Input | Supervised | GT type | Abs Rel | Sq Rel | RMSE | RMSE$_{log}$ | $\delta < 1.25$ | $\delta < 1.25^2$ | $\delta < 1.25^3$ |
|---|---|---|---|---|---|---|---|---|---|---|
| MonoDepth [19] | M→O | ✗ | Improved | 0.090 | 0.545 | 3.942 | 0.137 | 0.914 | 0.983 | 0.995 |
| PackNet-SfM [22] | M→O | ✗ | Improved | 0.071 | 0.359 | 3.153 | 0.109 | 0.944 | 0.990 | 0.997 |
| DRO (ours) | Multi | ✗ | Improved | 0.057 | 0.342 | 3.201 | 0.123 | 0.952 | 0.989 | 0.996 |
| Kuznietsov et al. [26] | One | ✓ | Improved | 0.089 | 0.478 | 3.610 | 0.138 | 0.906 | 0.980 | 0.995 |
| DORN [18] | One | ✓ | Improved | 0.072 | 0.307 | 2.727 | 0.120 | 0.932 | 0.984 | 0.995 |
| PackNet-SfM [22] | M→O | ✓ | Improved | 0.064 | 0.300 | 3.089 | 0.108 | 0.943 | 0.989 | 0.997 |
| BANet [33] | Multi | ✓ | Improved | 0.083 | – | 3.640 | 0.134 | – | – | – |
| DeepV2D (2-view) [34] | Multi | ✓ | Improved | 0.064 | 0.350 | 2.946 | 0.120 | 0.946 | 0.982 | 0.991 |
| DRO (ours) | Multi | ✓ | Improved | **0.047** | **0.199** | **2.629** | **0.082** | **0.970** | **0.994** | **0.998** |
| SfMLearner [50] | M→O | ✗ | Velodyne | 0.208 | 1.768 | 6.856 | 0.283 | 0.678 | 0.885 | 0.957 |
| CCNet [30] | M→O | ✗ | Velodyne | 0.140 | 1.070 | 5.326 | 0.217 | 0.826 | 0.941 | 0.975 |
| GLNet [10] | M→O | ✗ | Velodyne | 0.135 | 1.070 | 5.230 | 0.210 | 0.841 | 0.948 | 0.980 |
| MonoDepth [19] | M→O | ✗ | Velodyne | 0.115 | 0.882 | 4.701 | 0.190 | 0.879 | 0.961 | 0.982 |
| PackNet-SfM [22] | M→O | ✗ | Velodyne | 0.107 | 0.803 | 4.566 | 0.197 | 0.876 | 0.957 | 0.979 |
| DRO (ours) | Multi | ✗ | Velodyne | 0.088 | 0.797 | 4.464 | 0.212 | 0.899 | 0.959 | 0.980 |
| PackNet-SfM [22] | M→O | ✓ | Velodyne | 0.090 | 0.618 | 4.220 | 0.179 | 0.893 | 0.962 | 0.983 |
| DRO (ours) | Multi | ✓ | Velodyne | **0.073** | **0.528** | **3.888** | **0.163** | **0.924** | **0.969** | **0.984** |

Table 1: Quantitative results on the KITTI dataset. Eigen split is used for evaluation and seven widely used metrics are reported. Results on two ground-truth types are displayed since different methods are evaluated using different types. 'M→O' means multiple images are used in the training while one image is used for inference.

### 4.2. Datasets

**KITTI dataset.** The KITTI dataset is a widely used benchmark with outdoor scenes captured from a moving vehicle. We adopt the training/testing split proposed by Eigen et al. [14] with 22,600 training images and 697 testing images. There are two types of ground-truth depth. One is the original Velodyne Lidar points which are quite sparse. The other one is the improved annotated depth map, which uses five successive images to accumulate the Lidar points and stereo images to handle moving objects. For the improved depth type there are 652 images for testing.

**ScanNet dataset.** ScanNet [13] is a large indoor dataset with 1,513 RGB-D videos in 707 distinct environments. The raw data is captured from a depth camera. The depth maps and camera poses are obtained from RGB-D 3D reconstruction. We use the training/testing split proposed by [33] with 2,000 test image pairs from 90 scenes.

| Method | Supervised | Abs Rel | Sq Rel | RMSE | RMSE$_{log}$ | SI Inv | Rot (deg) | Tr (deg) | Tr (cm) | Time (s) |
|---|---|---|---|---|---|---|---|---|---|---|
| Photometric BA [16] | ✓ | 0.268 | 0.427 | 0.788 | 0.330 | 0.323 | 4.409 | 34.36 | 21.40 | – |
| DeMoN [37] | ✓ | 0.231 | 0.520 | 0.761 | 0.289 | 0.284 | 3.791 | 31.626 | 15.50 | – |
| BANet [33] | ✓ | 0.161 | 0.092 | 0.346 | 0.214 | 0.184 | 1.018 | 20.577 | 3.39 | – |
| DeepV2D (2-view) [34] | ✓ | 0.069 | 0.018 | 0.196 | 0.099 | 0.097 | 0.692 | 11.731 | 1.902 | 0.95 |
| DRO (ours) | ✗ | 0.140 | 0.127 | 0.496 | 0.212 | 0.210 | 0.691 | 11.702 | 1.647 | 0.11 |
| DRO (ours) | ✓ | **0.053** | **0.017** | **0.168** | **0.081** | **0.079** | **0.473** | **9.219** | **1.160** | 0.11 |

Table 2: Quantitative results on the ScanNet dataset. Five metrics of the depth and three metrics of the pose are reported.

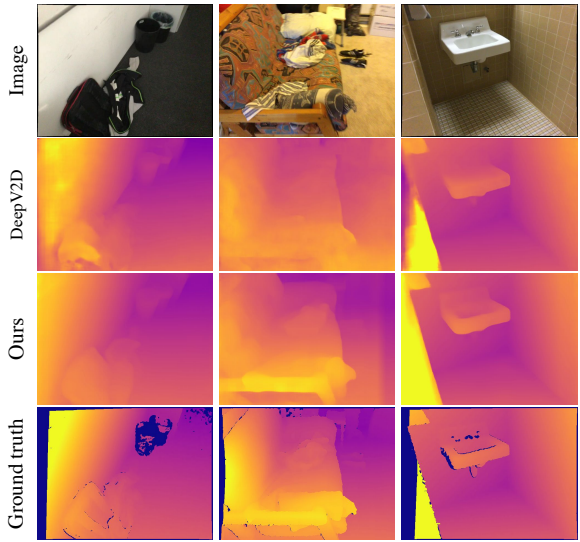

Figure 6: Qualitative results on the ScanNet dataset.

**SUN3D, RGB-D SLAM, and Scenes11.** SUN3D [42] contains indoor scenes with imperfect depths and poses estimated by SfM algorithms. RGB-D SLAM [32] provides high-quality camera poses obtained with an external motion-tracking system and noisy depth maps. Scenes11 [37] is a synthetic dataset generated from ShapeNet [8] with perfect depths and camera poses. For these datasets, we use the data processed and split by [37] for a fair comparison to previous methods, resulting in 29, 183 image sequences for training and 80 image sequences for testing in SUN3D, 3, 351 image sequences for training and 80 image sequences for testing in RGB-D SLAM, 17, 353 image sequences for training and 128 image sequences for testing in Scenes11.

### 4.3. Evaluation

**Evaluation on KITTI.** For outdoor scenes, we present the results of our method and some previous methods on the KITTI dataset in Table 1. State-of-the-art single-frame depth estimation methods [26, 18] and deep SfM methods [50, 30, 10, 19, 22, 33, 34] are listed. For a fair comparison, all SfM methods are evaluated under the two-view setting. From the results, our approach outperforms other methods by a large margin in both the supervised setting and the self-supervised setting. Also, the performance of our *self-supervised* model already surpasses most of previous *supervised* methods. The qualitative results of these outdoor scenes are shown in Figure 5, from which we can see our approach estimates better depth for distant and small-size or thin objects, e.g., people, motorbikes, and guideposts. Also, we predict sharper edges at object boundaries. Thin structures are usually recovered by fine updates in the last few iterations.

**Evaluation on ScanNet.** For indoor scenes, we evaluate our method on the ScanNet dataset in Table 2. For a fair comparison, all methods are evaluated under the two-view setting since there are only 2 images in the test split. The results of Photometric BA and DeMoN are cited from [33]. The results show that our model outperforms previous methods on both depth accuracy and pose accuracy. Our *self-supervised* model is able to predict the results that are comparable to previous *supervised* methods, especially on the pose accuracy. Among previous methods, DeepV2D performs best but it requires pre-training a complex pose solver first. Also, the inference time of their method is much longer than ours. Even using five views their performance is still not comparable to ours, with a larger depth error of 0.057. From the qualitative results shown in Figure 6, our model predicts the finer depth of the indoor objects and is robust in clutter.

**Evaluation on SUN3D, RGB-D SLAM, and Scenes11.** We further evaluate our approach on the SUN3D, RGB-D SLAM, and Scenes11 datasets using the data pre-processed by [37] for a fair comparison. In the experiments, however, we find some of the picked image sequences are not suitable for training an SfM model with little view overlap between neighboring images, which significantly degrades the performance of depth map estimation. Still, we achieve decent performance that is comparable to previous methods, especially in the pose accuracy, which is less affected by the insufficient view overlap.

### 4.4. Ablation Study

To better understand the individual components, we evaluate each module by an ablation study with the KITTI dataset and present the results in Table 4.

| Method | SUN3D | | | | | RGB-D | | | | | Scenes11 | | | | |
|---|---|---|---|---|---|---|---|---|---|---|---|---|---|---|---|
| | L1-inv | Sc-inv | L1-rel | Rot | Tran | L1-inv | Sc-inv | L1-rel | Rot | Tran | L1-inv | Sc-inv | L1-rel | Rot | Tran |
| DeMoN [37] | 0.019 | 0.114 | 0.172 | 1.801 | 18.811 | 0.028 | 0.130 | 0.212 | 2.641 | 20.585 | 0.019 | 0.315 | 0.248 | 0.809 | 8.918 |
| LS-Net [12] | 0.015 | 0.189 | 0.650 | 1.521 | 14.347 | 0.019 | 0.090 | 0.301 | **1.010** | 22.100 | 0.010 | 0.410 | 0.210 | 0.910 | 8.210 |
| BANet [33] | 0.015 | 0.110 | **0.060** | 1.729 | 13.260 | 0.008 | 0.087 | 0.050 | 2.459 | 14.90 | 0.080 | 0.210 | 0.130 | 1.298 | 10.370 |
| DeepSfM [41] | 0.013 | **0.093** | 0.072 | 1.704 | 13.107 | 0.011 | **0.071** | 0.126 | 1.862 | 14.570 | **0.007** | **0.112** | **0.064** | 0.403 | 5.828 |
| DRO (ours) | **0.011** | 0.108 | 0.076 | **1.334** | **10.988** | **0.004** | 0.087 | **0.046** | 2.839 | **11.390** | 0.010 | 0.167 | 0.096 | **0.366** | **3.705** |

Table 3: Quantitative results on the SUN3D, RGB-D SLAM, and Scenes11 datasets.

| Setting | | Abs Rel | Sq Rel | RMSE | $R_{log}$ | 1.25 | $1.25^2$ |
|---|---|---|---|---|---|---|---|
| w/o GRU | | 0.058 | 0.258 | 2.953 | 0.097 | 0.955 | 0.992 |
| w/o Alter | | 0.055 | 0.247 | 2.952 | 0.094 | 0.959 | 0.992 |
| w/o Cost | | 0.065 | 0.324 | 3.270 | 0.112 | 0.940 | 0.988 |
| Cost volume | | 0.049 | 0.214 | 2.804 | 0.086 | 0.966 | 0.994 |
| Full-setting | | 0.047 | 0.199 | 2.629 | 0.082 | 0.970 | 0.994 |
| Infer iterations | 0 | 0.094 | 0.529 | 4.014 | 0.150 | 0.891 | 0.974 |
| | 4 | 0.059 | 0.266 | 2.992 | 0.099 | 0.951 | 0.992 |
| | 8 | 0.049 | 0.208 | 2.687 | 0.084 | 0.968 | 0.994 |
| | 16 | 0.046 | 0.198 | 2.623 | 0.081 | 0.970 | 0.994 |
| | 24 | 0.046 | 0.199 | 2.626 | 0.082 | 0.970 | 0.994 |

Table 4: Ablation study on the KITTI dataset. The first six metrics of those used in Table 1 are reported here.

| Models | | Resolution | Memory | Time | Abs Rel |
|---|---|---|---|---|---|
| DeepV2D [34] | | $192 \times 1088$ | 6.75 G | 1.49 s | 0.064 |
| DRO | iterate 12 | $320 \times 960$ | 1.16 G | 0.12 s | 0.047 |
| | iterate 4 | $320 \times 960$ | 1.15 G | 0.049 s | 0.059 |

Table 5: Efficiency experiments on the KITTI dataset.

**GRU module.** The core module of our method is the recurrent optimizer. To see its effectiveness in optimization, we replace the GRU block with three convolutional layers. In the training, the depth error decreases to 0.058 in the first a few epochs but then the network diverges. This demonstrates that by leveraging the historical information in a recurrent optimizer, not only a superior optimum could be reached, but also the optimization would be more stable.

**Alternate update.** It is important to alternately update the depth and camera poses to decouple their influence in the feature-metric error. To verify this, we train a model where the optimizer predicts the updates for depth and camera poses simultaneously. The depth accuracy in this setting is almost as poor as the one without GRU. This proves the importance of the alternative optimization.

**Cost Volume.** One of the advantages of our method is that we do not need a heavy cost volume for optimization. Here, we replace the feature-metric cost map $C$ with a $H \times W \times 64$ cost volume. The cost volume is in a cascaded structure, i.e, the depth range of the volume is dynamically adjusted over iterations for better results. From the results shown in Table 4, the performance of using this heavy cost volume is similar to using the cost map, which proves that employing information in temporal domain can make up the lack of neighborhood information in spatial domain. Also, we test the performance of a model without the cost input. As expected, the error of the depth estimation is large since the optimizer loses the objective to minimize.

**Iteration times.** Until now, we only use 12 iterations of recurrent optimization for all experiments. We could also vary the number of iterations during inference. Here, we test different iteration numbers in the inference. Zero iteration means we do not update the initial depth and pose at all. According to the results in Table 4, our optimizer already outputs decent results after 4 iterations and predicts accurate results after 8 iterations, after which the depths are further refined with more iterations. This demonstrates that our optimizer has learned to optimize the estimation step by step. A model trained with a fixed number of iterations can be applied with more iterations in real applications to obtain finer results.

## 4.5. Run-time and Memory Efficiency

We compare our method with DeepV2D [34] in run-time and memory efficiency in Table 5. All experiments are conducted on the KITTI dataset with an Nvidia GTX 2080 Ti GPU. Compared to DeepV2D, our method achieves similar depth accuracy with only 4 iterations, which takes only about $1/6$ of GPU memory and $1/30$ of inference time comparing with DeepV2D. Even with 12 iterations, our method is still more than 10 times faster and reduces about $26\%$ depth errors. More efficiency experiments are shown in the supplements.

## 5. Conclusion

We propose a zeroth-order deep recurrent optimizer for addressing the structure-from-motion problem. Two gated recurrent units have been introduced to optimize scene structures and camera poses respectively. Our optimizer avoid computing a cost volume or gradients by exploiting temporal information during the optimization process. The experiments demonstrate that our method outperforms previous methods on both outdoor and indoor datasets, in both the supervised and self-supervised settings.

# DRO: Deep Recurrent Optimizer for Structure-from-Motion
# Appendix

## A. Network Structure

To illustrate the details of our framework, we display our network structure in Figure 1. The first module is the initialization block. Two base feature networks sharing the same parameters are employed to extract the features $\mathcal{F}_0$ and $\mathcal{F}_i$ of input images, which are leveraged to estimate the initial depth and pose. The feature networks are using ResNet18 [2] as the backbone, and the output heads are composed of two convolutional layers. For the pose head, a global average pooling layer is also adopted. Then, based on the current estimated depth, pose, and features $\mathcal{F}_0$ and $\mathcal{F}_i$, a cost $C_i$ is constructed. Afterward, the contextual networks use the same structure as the base feature networks to extract the contextual features $\mathcal{F}_0^c$ and $\mathcal{F}_i^c$, which are fed into the GRU modules. The GRU optimizers take the contextual features, the cost, and the current estimated depth or pose in, to estimate the variables increments.

## B. Time and Memory Consumption

One of the advantages of our approach compared to cost-volume-based methods is the efficiency in computation and memory consumption. To see how lightweight our optimizer is, we compare it in different numbers of iterations with DeepV2D [4], which is the method performing best before ours. We test the GPU memory occupancy and the network forward time of different numbers of iterations in the inference, as presented in Table 1. All experiments are conducted on the KITTI dataset with an Nvidia GTX 2080 Ti GPU.

We try two strategies for the model running different numbers of iterations. The first one is directly applying the model trained with 12 iterations and only adjusting the number of iterations in the inference. The second one is learning a new model from scratch with fewer iterations in the training, which is denoted with "*" in Table 1. The results show that the retrained models perform slightly better. Compared to DeepV2D, both strategies achieve better performance in four iterations, with much less GPU Memory and inference time. The model retrained from scratch with two iterations can even outperform DeepV2D. This allows our model to be applied to small-memory mobile devices and be employed

| Models | | Resolution | Memory | Time | Abs Rel |
|---|---|---|---|---|---|
| DeepV2D [4] | | $192 \times 1088$ | 6.75 G | 1.49 s | 0.064 |
| DRO | iterate 12 | $320 \times 960$ | 1.16 G | 0.12 s | 0.047 |
| | iterate 8 | $320 \times 960$ | 1.15 G | 0.086 s | 0.049 |
| | iterate 4 | $320 \times 960$ | 1.15 G | 0.049 s | 0.059 |
| | iterate 0 | $320 \times 960$ | 1.11 G | 0.011 s | 0.094 |
| DRO (retrained) | iterate 4* | $320 \times 960$ | 1.15 G | 0.049 s | 0.054 |
| | iterate 3* | $320 \times 960$ | 1.13 G | 0.042 s | 0.056 |
| | iterate 2* | $320 \times 960$ | 1.12 G | 0.032 s | 0.060 |
| | iterate 1* | $320 \times 960$ | 1.12 G | 0.022 s | 0.066 |
| | iterate 0* | $320 \times 960$ | 1.11 G | 0.011 s | 0.081 |

Table 1: Efficiency experiments on the KITTI dataset. Resolution of the input images, GPU Memory consumption, network forward time, and the relative absolute error of the depth map are reported. "*" denotes retrained from scratch.

in real-time applications.

## C. More Views

In an SfM task, the depths and poses could be estimated with arbitrary number of images. In previous experiments, however, only two images are used for inference for a fair comparison with other methods. Inference with more frames should predict better depths and poses, since there is more information provided. Hence, we feed five images into our model in the inference without retraining to see if it can perform better, as presented in Table 2. Compared to the results of using two views, the performance of using five views is better as expected.

## D. More Visualization Results

We display more qualitative results on the SUN3D dataset, RGB-D SLAM dataset, and Scenes11 dataset. The depth maps obtained with our supervised model and DeepSfM [5] are presented in Figure 2 and Figure 3, from which we can see that our model predicts smoother and cleaner depth maps.

In addition, our model can be trained in either the supervised setting or the self-supervised setting. The depth maps and poses estimated by our self-supervised model are already more accurate than that of many supervised models. To inspect how our self-supervised model performs, we compare the results of it with our supervised model and self-supervised PackNet-SfM [1] in Figure 4. Compared to PackNet-SfM, our model predicts better details of small-size and distinct objects, and estimates correct depth of the hard objects like the white guideboard. Compared to the supervised model, there is still a gap but the gap is not large. Sometimes the self-supervised model can estimate better depth than the supervised model.

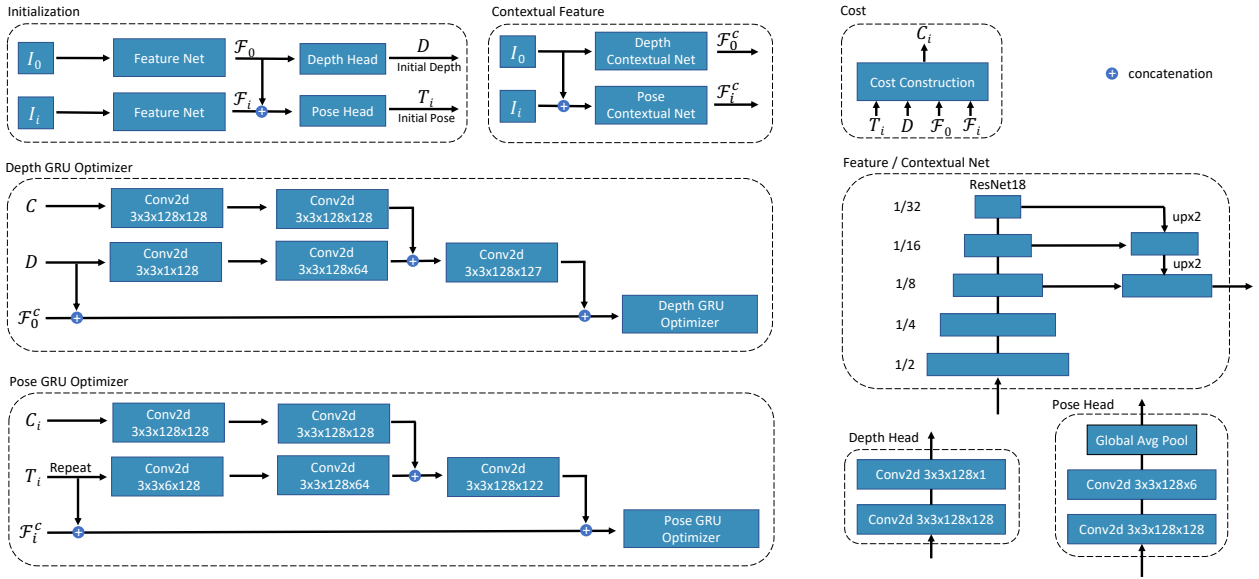

Figure 1: The details of modules in our framework. The reference image $I_0$ and neighbor image $I_i$ are fed into the base feature networks to extract the features $\mathcal{F}_0$ and $\mathcal{F}_i$, which are then input to the depth head and pose head to predict the initial depth and initial pose. Afterward, the contextual networks extract the contextual features $\mathcal{F}_0^c$ and $\mathcal{F}_i^c$, which are fed into the GRU optimizers together with the depth, pose, and cost. The details of the cost construction, feature networks, and output heads are also presented.

| Method | #View | Abs Rel | Sq Rel | RMSE | RMSE$_{log}$ | SI Inv | Rot (deg) | Tr (deg) | Tr (cm) |
|---|---|---|---|---|---|---|---|---|---|
| BANet [3] | 2 | 0.161 | 0.092 | 0.346 | 0.214 | 0.184 | 1.018 | 20.577 | 3.390 |
| BANet [3] | 5 | 0.091 | 0.058 | 0.223 | 0.147 | 0.137 | 1.009 | 14.626 | 2.365 |
| DeepV2D [4] | 2 | 0.069 | 0.018 | 0.196 | 0.099 | 0.097 | 0.692 | 11.731 | 1.536 |
| DeepV2D [4] | 5 | 0.057 | **0.010** | 0.168 | 0.080 | 0.077 | 0.628 | 10.800 | 1.373 |
| DRO (ours) | 2 | 0.053 | 0.017 | 0.168 | 0.081 | 0.079 | 0.473 | 9.219 | 1.160 |
| DRO (ours) | 5 | **0.049** | 0.015 | **0.157** | **0.075** | **0.074** | **0.463** | **8.871** | **1.139** |

Table 2: Quantitative results of using five views on the ScanNet dataset. Five metrics of the depth and three metrics of the pose are reported.

## References

[1] Vitor Guizilini, Rares Ambrus, Sudeep Pillai, Allan Raventos, and Adrien Gaidon. 3d packing for self-supervised monocular depth estimation. In *Proceedings of the IEEE Conference on Computer Vision and Pattern Recognition*, pages 2485–2494, 2020. 2

[2] Kaiming He, Xiangyu Zhang, Shaoqing Ren, and Jian Sun. Deep residual learning for image recognition. In *Proceedings of the IEEE Conference on Computer Vision and Pattern Recognition*, pages 770–778, 2016. 1

[3] Chengzhou Tang and Ping Tan. Ba-net: Dense bundle adjustment networks. In *Proceedings of the International Conference on Learning Representations*, 2019. 2

[4] Zachary Teed and Jia Deng. Deepv2d: Video to depth with differentiable structure from motion. In *Proceedings of the International Conference on Learning Representations*, 2020. 1, 2

[5] Xingkui Wei, Yinda Zhang, Zhuwen Li, Yanwei Fu, and Xiangyang Xue. Deepsfm: Structure from motion via deep bundle adjustment. In *Proceedings of the European Conference on Computer Vision*, pages 230–247. Springer, 2020. 1

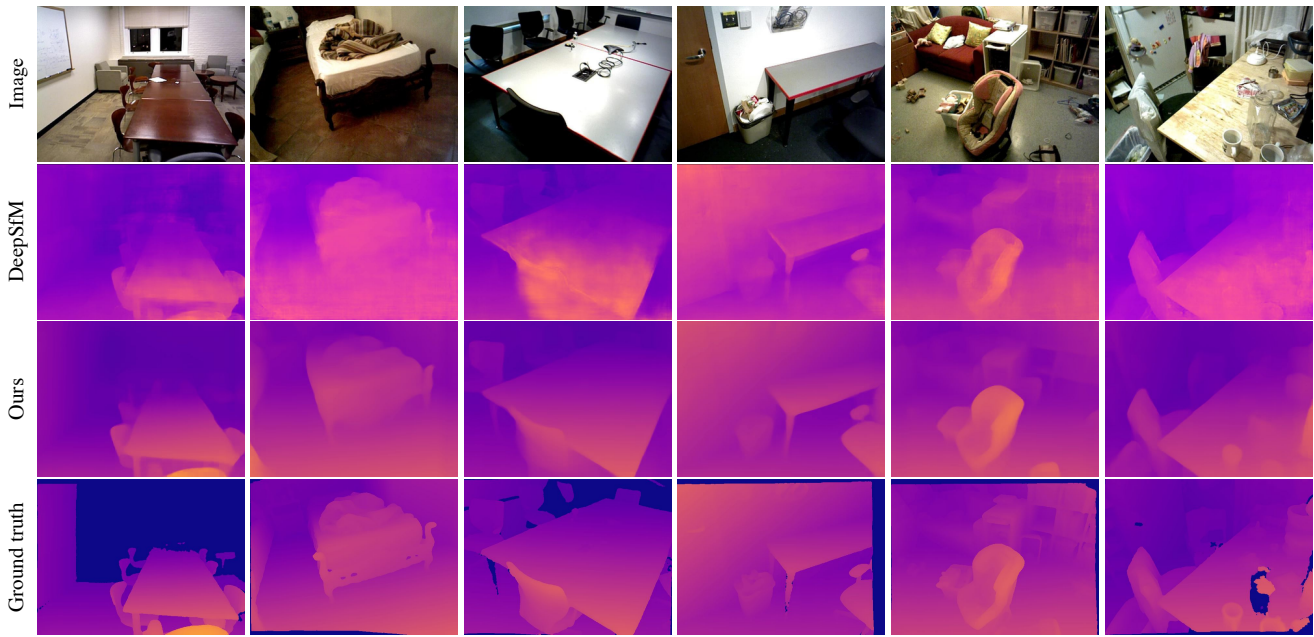

Figure 2: Qualitative results of on the SUN3D dataset. The depth maps obtained by DeepSfM and our approach are displayed with the input images and ground-truth depth maps.

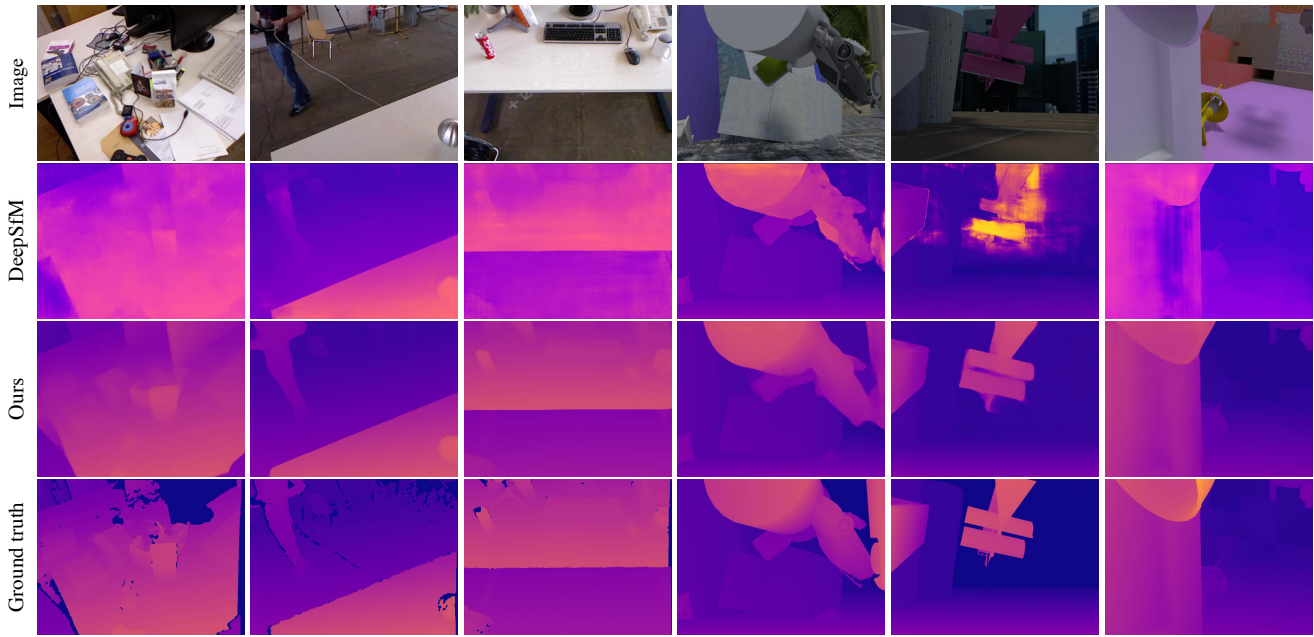

Figure 3: Qualitative results on the RGB-D and Scenes11 dataset. The left three columns are results of RGB-D images and the right three are those of Scenes11 images. The depth maps obtained by DeepSfM and our approach are displayed with the input images and ground-truth depth maps.

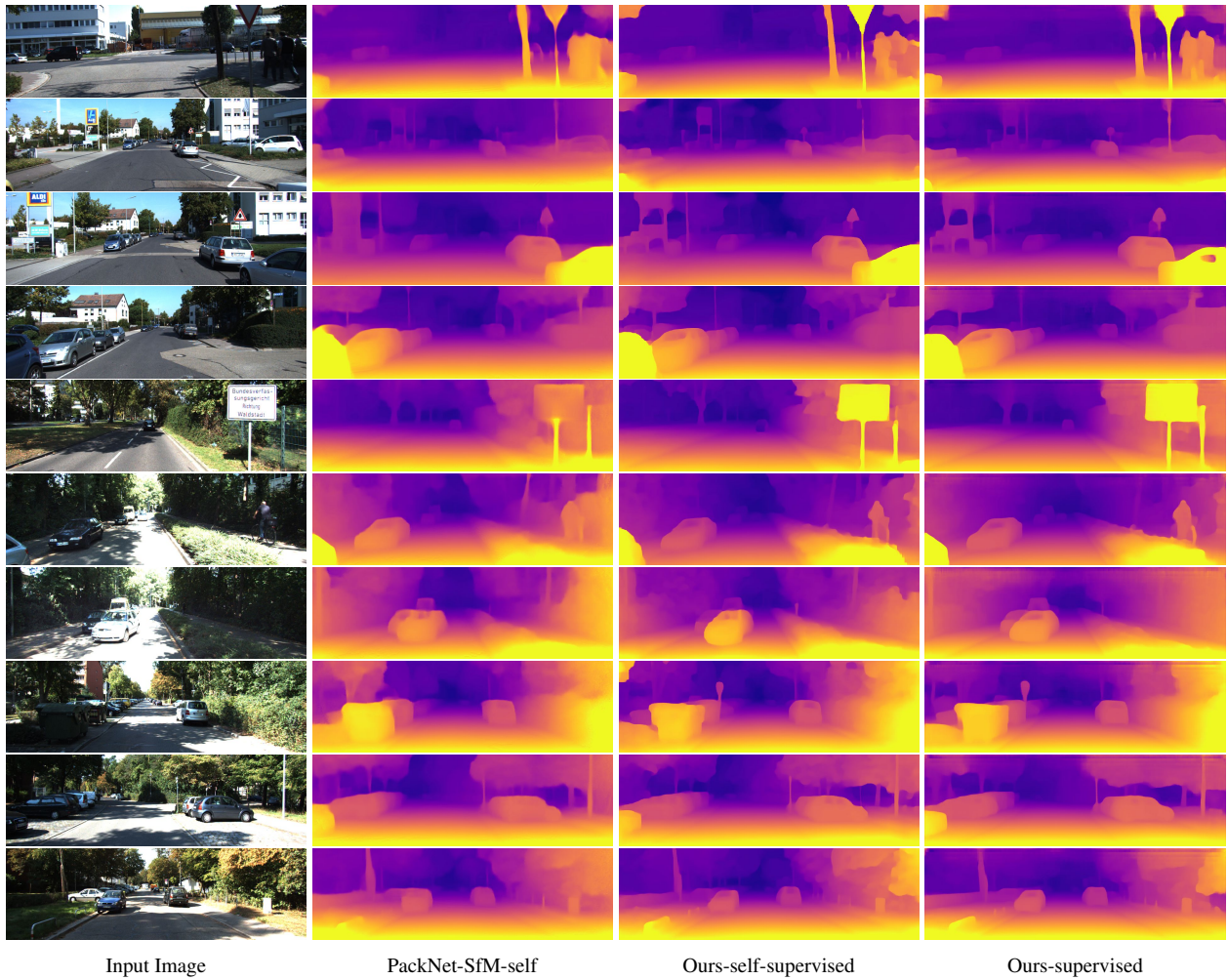

| Input Image | PackNet-SfM-self | Ours-self-supervised | Ours-supervised |

Figure 4: Qualitative results of self-supervised models and our supervised model on the KITTI dataset. "PackNet-SfM-self" is the model of PackNet-SfM trained in their self-supervised setting.